\pdfoutput=1
\documentclass[11pt]{article}

\usepackage{acl}

\usepackage{times}
\usepackage{latexsym}

\usepackage[T1]{fontenc}
\usepackage[utf8]{inputenc}

\usepackage{microtype}

\usepackage{inconsolata}

\usepackage{booktabs}
\usepackage{amssymb}
\usepackage{amsmath}
\usepackage{multirow}
\usepackage{graphicx}
\usepackage{makecell}
\usepackage{longtable}

\newcommand{\mocl}{MoCL}
\newcommand{\blue}[1]{\textcolor{black}{#1}}

\title{Rehearsal-Free Modular and Compositional Continual Learning \\ for Language Models}

\author{
Mingyang Wang$^{1,2}$ \hspace*{0.2cm}
{Heike Adel$^{3}$} \hspace*{0.2cm} 
{Lukas Lange$^{1}$} \\
 {\bf Jannik Str\"{o}tgen$^{4}$ \hspace*{0.2cm} Hinrich Sch\"{u}tze$^{2}$} \\
  $^1$Bosch Center for Artificial Intelligence, Renningen, Germany \\
  $^2$LMU Munich, Germany \hspace*{0.2cm}
  $^3$Hochschule der Medien, Stuttgart, Germany \\
  $^4$Karlsruhe University of Applied Sciences, Germany \\
  \texttt{mingyang.wang2@de.bosch.com} 
  }

\begin{document}
\maketitle

\begin{abstract}
Continual learning aims at incrementally acquiring new knowledge while not forgetting existing knowledge. To overcome catastrophic forgetting, methods are either rehearsal-based, i.e., store data examples from previous tasks for data replay, or isolate parameters dedicated to each task. However, rehearsal-based methods raise privacy and memory issues, and parameter-isolation continual learning does not consider interaction between tasks, thus hindering knowledge transfer. In this work, we propose \mocl, a rehearsal-free \textbf{Mo}dular and \textbf{C}ompositional Continual \textbf{L}earning framework which continually adds new modules to language models and composes them with existing modules.
Experiments on various benchmarks show that \mocl\  outperforms state of the art and effectively facilitates knowledge transfer.

% without suffering from catastrophic forgetting.

\end{abstract}
% \newpage

\begin{figure*}[htbp]
    \centering
    \includegraphics[width=1\linewidth]{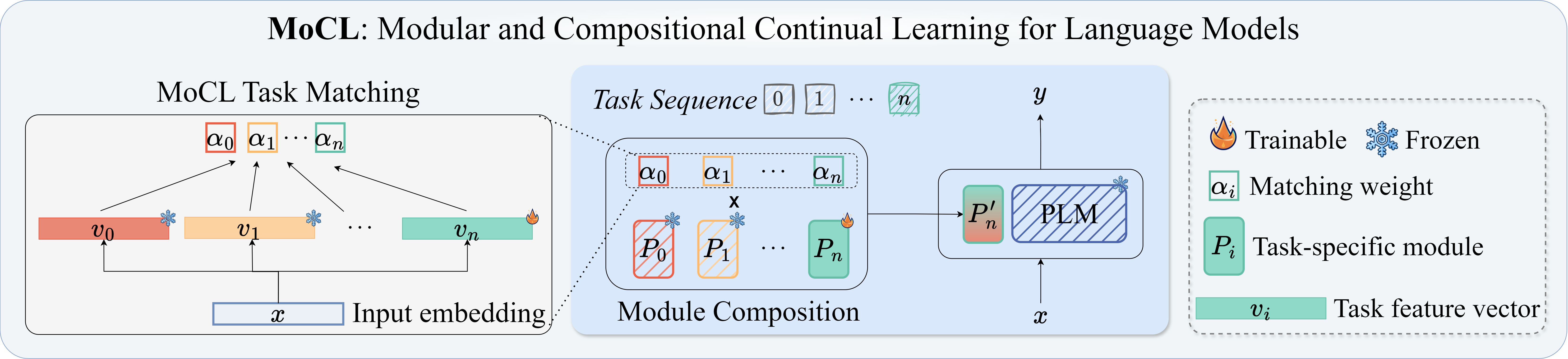}
    \caption{Overview of the \mocl\ framework for continual learning. \mocl\ continually adds new modules to language models and composes existing and new modules based on task matching weights for learning the new task. }
    \label{fig:main-figure}
\end{figure*}

\section{Introduction}

To effectively deploy machine learning (ML) models in real-world settings, they need to adopt \textit{continual learning} (CL), i.e., incrementally acquire, update and accumulate knowledge to evolve continually and stay effective over time \cite{chen2018continual}. %given real-world dynamics. 
%This ability, often referred to as \textit{continual learning} (CL), is essential for ML models to remain effective over time. 
However, CL often suffers from \textit{catastrophic forgetting} \cite{mccloskey1989catastrophic}: The knowledge learned at early stages of training is overwritten by subsequent model updates.

A commonly used strategy to mitigate catastrophic forgetting is to store training samples from prior tasks along the continual learning process and train the model jointly with samples from prior and current tasks (\textit{rehearsal}) \cite{rebuffi2017icarl}. However, training samples of prior tasks are not always available due to storage or privacy constraints \cite{wang2023comprehensive}.

Another line of work allocates task-specific parameters to overcome catastrophic forgetting, often referred to as \textit{parameter isolation-based} CL. %The inter-task interference, a main reason for catastrophic forgetting \cite{wang2023comprehensive}, can therefore be alleviated.
Although inter-task interference leads to catastrophic forgetting \cite{wang2023comprehensive}, knowledge transfer across tasks could be promising. However, those approaches do not enable effective knowledge transfer. Recent parameter isolation-based methods either separately train task-specific modules, completely excluding knowledge transfer \cite{wang-etal-2023-rehearsal}, or progressively concatenate all previous task-specific modules with the current task module \cite{razdaibiedina2022progressive}, without considering if the interaction between tasks is
``positive'' (knowledge transfer boosting performance) or ``negative'' (knowledge interference hurting performance).

To address these challenges, we introduce \mocl, a \textbf{Mo}dular and \textbf{C}ompositional Continual \textbf{L}earning framework for language models.
% Here is our contribution
\mocl\ avoids catastrophic forgetting without storing additional data and facilitates effective knowledge transfer via module composition.
Specifically, \mocl\ allocates task-specific parameters using parameter-efficient fine-tuning (PEFT) modules.\footnote{\blue{We use two PEFT methods, prefix-tuning \cite{li2021prefix} and LoRA \cite{hu2021lora} in this work to be consistent with prior works (see  Section~\ref{sec:training-details} for details). Other PEFT methods, such as Adapter \cite{houlsby2019parameter}, can also be combined with MoCL in general. We leave such exploration for future work.}}
During training, \mocl\ continually adds new task-specific modules to language models. To avoid catastrophic forgetting, the task-specific module is frozen once the training on the respective task is finished. Additionally, \mocl\ facilitates knowledge transfer across tasks by composing existing and new modules based on task matching weights while learning the new task. 
In our evaluation on \textit{near-domain} and \textit{far-domain} continual learning benchmarks, 
% (Web-Of-Science \cite{kowsari2017hdltex}, Afrisenti \cite{muhammad2023afrisenti} and MTL5 \cite{de2019episodic}), 
\mocl\ outperforms state-of-the-art methods under the task-incremental learning setting where the task identities are available during testing.
%We evaluate \mocl\ on \textit{near-domain} and \textit{far-domain} continual learning benchmarks (Web-Of-Science \cite{}, Afrisenti \cite{} and MTL5 \cite{}). On all benchmarks, \mocl\ outperforms the state-of-the-art methods under the task-incremental learning setting where the task identities are available during testing. 
It further demonstrates strong abilities to transfer knowledge of previous tasks to the new tasks. Furthermore, the task matching strategy of \mocl\ enables task composition during testing. As a result, \mocl\ effectively addresses the continual learning problem in the challenging class-incremental setting where task identities are not provided during testing.

The code base for MoCL is available online.\footnote{https://github.com/boschresearch/MoCL-NAACL-2024}

% under task-incremental and class-incremental settings, where the task identities are or are not available during testing, respectively. On all benchmarks, \mocl\ outperforms the state-of-the-art methods and demonstrates the ability to transfer knowledge of previous tasks to the new task.

% Furthermore, the task matching strategy of \mocl\ enables task composition during testing. As a result, \mocl\ shows better or competitive performance in the challenging class-incremental setting where task identities are not provided during testing.

% \begin{figure}
% \centering
%     \includegraphics[width=.45\textwidth]{figures/Picture6.png}
    
%     \caption{Comparison between MCCL (ours) and previous parameter isolation-based continual learning methods. Unlike prior work which simply isolates task-specific modules or concatenates all task-specific modules while continual learning, we take the task similarity into consideration, based on which modules are composed for efficient new task learning and better knowledge transfer.
%     }
%     \label{fig:teaser}
% \end{figure}

\section{Related Work}

In line with previous work \cite{de2021continual, ke2022continual, wang2023comprehensive}, we group CL strategies into three categories. (i) \textit{Regularization}-based methods add explicit regularization terms to preserve the knowledge of previous tasks \cite{li2017learning, kirkpatrick2017overcoming, aljundi2018memory}. As regularizing knowledge tends to have suboptimal performance, it is often used in combination with other methods. 
(ii) \textit{Rehearsal}-based methods address catastrophic forgetting by saving old training samples in a memory buffer \cite{rebuffi2017icarl, rolnick2019experience, zhang-etal-2022-continual}, or training generative models to provide pseudo samples of previous tasks \cite{shin2017continual, su2019generative} for future rehearsal. (iii) \textit{Parameter isolation}-based methods assign isolated parameters dedicated to each task along the CL process to prevent interference between tasks \cite{madotto2020continual, zhang2022continual, razdaibiedina2022progressive, wang-etal-2023-rehearsal, wang2023orthogonal}.
 
%While rehearsal-based methods have shown strong results for different CL tasks, storing training data raises data privacy concerns and causes larger memory cost. 
Since rehearsal-based methods raise memory and data privacy issues, we
%The training process of generation models is also computationally expensive. 
%This motivates us to 
focus on rehearsal-free CL methods. %to mitigate catastrophic forgetting. 
\mocl\ falls into the category of parameter isolation-based continual learning, i.e., we allocate task-specific parameters 
%so that the incremental tasks can be learned in a (partially) separated way 
to avoid knowledge interference. In contrast to related work, we additionally encourage knowledge transfer considering the relatedness across tasks.

\section{Continual Learning Basics / Notation}
\label{sec:cl-basics}
In this work, we focus on continual learning (CL) on a sequence of text classification tasks. Specifically, we denote the sequence of tasks as $\{T_1, \dots, T_N\}$. 
%The $ n^\text{th}$ 
Each task $T_n$ contains a set of input samples $ \{(x^{i}_{n}, y^{i}_{n})\}$, where $x^{i}_{n}$ is the input text, $y^{i}_{n}$ is the ground-truth label, and $n  \in \{1, \dots, N\}$ is the task identity. A CL model aims to solve the series of tasks which arrive sequentially. The overarching goal is to optimize the model’s average performance across all tasks after learning them in the sequence. As we focus on rehearsal-free continual learning, data from earlier tasks is not available when training later tasks, i.e., our model does not suffer from the aforementioned shortcomings of rehearsal-based methods, such as memory issues. 

While in many benchmark settings, the task identity $n$ is provided, it is not a realistic assumption that task identities are available in real-world setups. Thus, we consider two setups: task-incremental learning (TIL) and class-incremental learning (CIL). In TIL, the task identities are available in both training and testing. In CIL, the task identities are only provided during training.% Under realistic constraints, the task identity $n$ might not always be available. We consider the task-incremental learning (TIL) and class-incremental learning (CIL) scenarios where (1) TIL: the task identities are available in both training and testing, and (2) CIL: the task identities are only provided in training.
\footnote{For better readability, we also refer to the domain-incremental learning (DIL), where tasks have the same label space but different input distributions, with and without test-time task identities as CIL and TIL, respectively; see Appendix~\ref{sec:cl-setting-details} for a more rigorous definition.}

\section{Method}
\label{sec:method}

We propose \mocl, a novel CL approach for language models to tackle catastrophic forgetting and enhance knowledge transfer at the same time.
% Motivated by prior parameter-efficient CL work \cite{razdaibiedina2022progressive, wang-etal-2023-rehearsal}, 

%\paragraph
\noindent \textbf{Avoiding Catastrophic Forgetting.}
We utilize \blue{two representative PEFT methods, prefix-tuning \cite{li2021prefix} and LoRA \cite{hu2021lora}} for allocating task-specific parameters to LMs, avoiding catastrophic forgetting without storing data samples.
In particular, \mocl\ adds a set of trainable PEFT parameters (\blue{prefix or LoRA}) to the frozen pretrained language model (PLM) for downstream task fine-tuning. Instead of updating the whole model, only a small number of the \blue{PEFT} parameters are trained.
As illustrated in Figure \ref{fig:main-figure}, \mocl\ optimizes the task-specific modules and keeps the PLM frozen. For each task $T_n$ in the sequence,  we initialize a trainable module $P_n$ for fine-tuning.
After the training on one task is finished, the corresponding PEFT parameters are frozen to preserve the task-specific knowledge in the following training process, thus avoiding catastrophic forgetting.

%\paragraph
\noindent \textbf{Enabling Knowledge Transfer.}
\mocl\ introduces task feature vectors for task matching and composes old and new modules for learning. This composition strategy facilitates effective knowledge transfer, which is often ignored by prior work. 
In particular, while learning on $T_n$, the previously acquired knowledge, which is encoded in the respective PEFT module $(P_1, \dots, P_{n-1})$, is reused via a weighted summation, denoted as $P'_n = \sum_{k=1}^{n} \alpha_k P_{k}$. Here, $P_{k}$ is the module specific to the $k^\text{th}$ task and $\alpha_k$ is the weight determining the contribution of $P_{k}$ for new task learning. We detail its computation below. Finally, the composed module $P'_n$ is combined with the PLM, consisting of all the module components up to the current task.
 
 To calculate the contribution weights $\alpha_k$ of each task-specific module, we introduce trainable task feature vectors $V \in \mathbb R^{N \times D}$ to capture salient features of tasks in the CL sequence. Note that each task-specific vector $v \in \mathbb R^D $ has the same dimension as the input embeddings $x_n \in \mathbb R^D$ (i.e., the embeddings from the PLM encoder). Then, we calculate the cosine similarity between the input embeddings $x_n$ and feature vectors up to the current $n^\text{th}$ task $V\left[:n\right]$ as task matching scores $\alpha\left[:n\right] = \cos (x_n, V\left[:n\right])$.

%\paragraph
\noindent \textbf{Training and Inference.}
The training objective for the $n^\text{th}$ task is to find the PEFT module $P_{n}$ and the task feature vector $v_n$ that  minimize the cross-entropy loss of training examples, and, at the same time, maximize the cosine similarity between
% the task-specific vector 
$v_n$  and the corresponding task input embeddings $x_n$:

\begin{small}
\begin{equation}
\min_{P_n, v_n} - \sum_{x_n, y_n} \log p(y_n | x_n, P'_{n}, \theta) - \sum_{x_n} \cos (x_n, v_n)
\end{equation}
\end{small}

During inference, as the task identities are available in the TIL setting, we directly select the task-specific module for inference. In the CIL setting, we use the matching scores 
% $\{ \alpha_i\}_{i=1}^{N}$ 
between task inputs and the feature vectors for module composition. The resulting module is combined with the PLM for inference.

\section{Experimental Setup}
In this section, we describe our experimental setup.

\subsection{Datasets}

\begin{table}[htbp]
  \centering
  \scalebox{0.8}{
    \begin{tabular}{llll}
    \toprule
    \textbf{Dataset} & \textbf{Class} & \textbf{Task Type} & \textbf{Domain} \\
    \midrule
    AGNews & 4     & Topic classification & News \\
    Yelp  & 5     & Sentiment anlysis & Yelp reviews \\
    Amazon & 5     & Sentiment anlysis & Amazon reviews \\
    DBPedia & 14    & Topic classification & Wikipedia \\
    Yahoo & 10    & Q\&A  & Yahoo Q\&A \\
    \bottomrule
    \end{tabular}%
    }
  \caption{\blue{Details of the MTL5 dataset.}}
  \label{tab:mtl5-dataset}%
\end{table}%

Following \newcite{wang-etal-2023-rehearsal}, we distinguish benchmarks according to the domain similarity of tasks.
%use \textit{near-domain} and \textit{far-domain} CL benchmarks according to the domain relevance between tasks. %In the \textit{near-domain} benchmark, tasks are related to some extent, while \textit{far-domain} tasks are more distinct.
As \textit{near-domain} benchmarks, we use the Web-of-Science (WOS) document classification dataset \cite{kowsari2017hdltex} consisting of 7 tasks, and AfriSenti \cite{muhammad2023afrisenti}, a multilingual sentiment analysis dataset with 12 African languages. As  \textit{far-domain} benchmark, we use the widely adopted MTL5 dataset \cite{de2019episodic}, including 5 text classification tasks. \blue{We summarize the details of MTL5 in Table \ref{tab:mtl5-dataset}.}
% On all aforementioned datasets, we assume tasks have disjoint label spaces, i.e., their classification heads are different if applied, although the Afrisenti tasks share the same label space. 
We adopt the same multiple task orders as the prior works for evaluation. Detailed task information is provided in Appendix~\ref{sec:dataset-info}.

% AG News (news classification), Amazon reviews (sentiment analysis), DBPedia (text classification), Yahoo Answers (Q\&A classification) and Yelp reviews (sentiment analysis).

% with disjoint label space for each task. Additionally, we also experiment on AfriSenti, a multilingual sentiment analysis dataset to evaluate the capability of CL algorithms to incrementally learn tasks in different languages. We consider it as a near-domain benchmark as the languages included in AfriSenti share some similarities and the tasks share the same label space. 

\begin{table}%[htbp]
  \centering
  \scalebox{0.8}{
    \begin{tabular}{l|c|cccc}
    \toprule
    \multicolumn{1}{c|}{\multirow{2}[2]{*}{\textbf{Method}}} & \multirow{2}[2]{*}{\textbf{WOS}} & \multicolumn{4}{c}{\textbf{AfriSenti Orders}} \\
          &       & \textbf{AVG} & \textbf{1} & \textbf{2} & \textbf{3} \\
    \midrule
    \blue{Sequential FT-F} & \blue{47.15} & \blue{6.17} & \blue{5.62} & \blue{6.52} & \blue{6.30}\\
    Sequential FT-P & 53.86 & 49.10 & 50.05 & 49.74 & 47.53 \\
    Per-task FT & 82.78 & 52.41 & 52.41 & 52.41 & 52.41 \\
    % Replay$^\dagger$ & 77.86 & 46.86 & 47.29 & 48.60 & 44.69 \\
    % LwF$^\dagger$   & 30.96 & 6.81 & 5.40 & 8.21 & 6.81 \\
    % Replay+LwF$^\dagger$ & 76.78 & 37.76 & 33.31 & 42.36 & 37.62 \\
    % L2P-R & 77.10 &       &       &       &  \\
    ProgPrompt & 89.93 & 49.07 & 50.16 & 46.74 & 50.30 \\
    EPI & 77.83 & 43.10 & 41.49 & 42.65 & 45.16\\
    % EPI (TIL) & 82.78 &       &       &       &  \\
    % \textbf{Ours} (CIL) & 79.23 &       &       &       &  \\
    \textbf{\mocl\ (Ours)}& \textbf{90.59} & \textbf{56.77} & \textbf{57.05} & \textbf{56.52} & \textbf{56.74} \\
    \bottomrule
    \end{tabular}%
    }
  \caption{TIL results on near-domain WOS and AfriSenti datasets. \blue{\mocl\ outperforms existing continual learning methods on both datasets, suggesting \mocl\ effectively facilitates knowledge transfer across near-domain tasks.}
  }
  \label{tab:res-til-near}%
\end{table}%

\subsection{Training Details}
\label{sec:training-details}
We adopt four LMs for these datasets in line with prior works \cite{razdaibiedina2022progressive, wang2023orthogonal, wang-etal-2023-rehearsal}.
%\footnote{In general, \mocl\ is compatible with any transformer-based model.}
We use encoder-based models for WOS, AfriSenti and MTL5 datasets (BERT \cite{devlin2018bert}, AfroXLMR \cite{alabi2022adapting} and BERT, respectively), the encoder-decoder T5 model \cite{raffel2020exploring} \blue{as well as the decoder-based Llama 2-7B model \cite{touvron2023llama}} for MTL5 under the few-shot setting. \blue{For all models except Llama 2, we use prefix-tuning as the task-specific modules, and LoRA as the task modules on Llama 2. All design choices are kept consistent with previous works to ensure a fair comparison.} \blue{The reported results are the average performance after training on all tasks consecutively.} All results are averaged over three random seeds. The detailed experimental settings are provided in Appendix~\ref{sec:implement-details}.

\subsection{Baselines}
\label{sec:baselines}
To compare different CL methods, we include the following baselines\footnote{\blue{For consistency, we include the results of baseline methods compatible with multiple base models used in this work. Results of other baselines which are specifically designed for certain LMs can be found in Appendix~\ref{sec:res-details}.}}:
Sequential fine-tuning continuously fine-tunes the language model on the task sequence: \blue{\textbf{Sequential FT-F} means all model parameters are updated (fully fine-tuning),\footnote{\blue{We did not run the Sequential FT-F experiments on Llama 2 because of the computational overhead and its poor performance in other experimental setups.}} while \textbf{Sequential FT-P} only fine-tunes the PEFT parameters}; \textbf{Per-task FT} trains a separate PEFT module for each task; and the parameter isolation-based methods \textbf{ProgPrompt} \cite{razdaibiedina2022progressive}, \textbf{EPI} \cite{wang-etal-2023-rehearsal} and \blue{O-LoRA \cite{wang2023orthogonal}}. A detailed description of these methods can be found in Appendix~\ref{sec:baseline-details}.

\section{Experimental Results}
\label{sec:exp-res}
In this section, we discuss our experimental results.
%the experimental results of \mocl\ and baseline methods on the aforementioned continual learning benchmarks. % We first focus on the task-incremental learning setting in Section \ref{sec:res-near-domain} and \ref{sec:res-far-domain}, then we present the class-incremental learning results in Section \ref{sec:res-cil} . Finally we compare \mocl\ with prior work from the forward transfer perspective in Section \ref{sec:res-fwt}.

% \subsection{Near-Domain Benchmark}
% \label{sec:res-near-domain}
\subsection{\mocl\ for Task-Incremental Learning}
\label{sec:res-til}
%\paragraph
\noindent \textbf{Near-domain.} 
As shown in Table \ref{tab:res-til-near}, \mocl\ outperforms state-of-the-art methods on both benchmarks. 
It is 
%Comparing \mocl\ with per-task fine-tuning, our approach is 
7.81 and 4.36 points better than training each task with an individual model (per-task FT), indicating it realizes effective knowledge transfer. % across tasks in the continual learning process.

Since EPI consists of task identification and per-task fine-tuning, its performance depends on the task identification accuracy.
%The prior work EPI learns each task individually and is additionally equipped with a task identification module for the CIL setting. The EPI results in Table \ref{tab:res-til-near} and \ref{tab:res-til-far} are both based on the CIL setting.\footnote{The TIL results of EPI are identical to per-task fine-tuning.} Therefore, the performance of EPI highly depends on the task identification accuracy. 
While it achieves comparable results with per-task fine-tuning on WOS, the performance degrades on AfriSenti, where different languages could be harder to differentiate. %similar, and it is hard to differentiate these languages based on their task identification module. %purely based on their sentence embeddings.

While \mocl\ achieves comparable results to ProgPrompt on WOS (0.66 percentage points better), the performance gap on AfriSenti is considerably higher (7.7 points better).
%ProgPrompt achieves comparable performance on WOS to our approach (0.66 percentage points lower than \mocl), the performance gap on AfriSenti is considerably higher (ProgPrompt is 7.7 points lower).  
We assume this is due to the suboptimal knowledge transfer of ProgPrompt, which we will analyze in Section \ref{sec:res-fwt}.

% \subsection{Far-Domain Benchmark}
% \label{sec:res-far-domain}
%\paragraph
\noindent \textbf{Far-domain.} Table \ref{tab:res-til-far} provides the results on MTL5 using BERT%(encoder model)
, T5 %(encoder-decoder model)
and Llama 2 models% (decoder model)
. \mocl\ again outperforms other CL methods in both cases across different task orders. Its advantage over per-task fine-tuning is less pronounced, which is due to the fact that far-domain tasks share weaker similarities. %, limiting the knowledge transfer. 
%Nevertheless, \mocl\ still consistently outperforms other continual learning methods.

\begin{table*}[h]
  \centering
  \scalebox{0.82}{
    \setlength\tabcolsep{4pt}
    \begin{tabular}{l|ccccc|cccc}
    \toprule
    \multicolumn{1}{c|}{\multirow{2}[2]{*}{\textbf{Method}}} & \multicolumn{5}{c|}{\textbf{MTL5 (BERT) Orders}}  & \multicolumn{4}{c}{\textbf{MTL5 (T5) Orders}} \\
          & \textbf{AVG} & \textbf{1} & \textbf{2} & \textbf{3} & \textbf{4} & \textbf{AVG} & \textbf{1} & \textbf{2} & \textbf{3} \\
    \midrule
    \blue{Sequential FT-F} & \blue{14.8}  & \blue{27.8}  & \blue{26.7}  & \blue{4.5}   & \blue{18.4} & \blue{28.5}  & \blue{18.9}  & \blue{24.9}  & \blue{41.7} \\
    Sequential FT-P & 66.5  & 66.4  & 65.7  & 65.4   & 68.5 & 27.2  & 24.6  & 30.3 & 25.0 \\
    Per-task FT & 79.0  & 79.0  & 79.0  & 79.0  & 79.0 & 75.1  & 75.1 & 75.1 & 75.1 \\
    % Replay $^\diamond$  & 57.8  & 67.2  & 64.7  & 64.7  & 44.6 \\
    % MBPA++ $^\diamond$ & 70.6  & 70.8  & 70.9  & 70.2  & 70.7 \\
    % IDBR $^\diamond$ & 76.3  & 75.9  & 76.2  & 76.4  & 76.7 \\
    ProgPrompt $^\diamond$ & 77.9  & 78.0  & 77.9  & 77.9  & 77.9 & 75.1  & 75.0  & 75.0  & 75.1 \\
    EPI $^\dagger$ & 77.3  & 77.4  & 77.3  & 77.2  & 77.4 & 56.4 & 49.7 & 54.1 & 65.3\\
    % EPI (TIL) & 79.0  & 79.1  & 79.0  & 78.9  & 78.9 \\
    % \textbf{Ours} (CIL) & \red{74.1} & \red{73.0} & \red{74.0} & \red{75.8} & \red{73.6} \\
    \textbf{\mocl}\ \textbf{(Ours)} & \textbf{79.4} & \textbf{79.3} & \textbf{79.6} & \textbf{79.2} & \textbf{79.4} & \textbf{75.9} & \textbf{75.6} & \textbf{75.4} & \textbf{76.7} \\
    \bottomrule
    \end{tabular}}
  \scalebox{0.82}{
    \setlength\tabcolsep{4pt}
    \begin{tabular}{l|cccc}
    \toprule
    \multicolumn{1}{c|}{\multirow{2}[2]{*}{\blue{\textbf{Method}}}} & \multicolumn{4}{c}{\blue{\textbf{MTL5 (Llama 2) Orders}}} \\
          & \textbf{AVG} & \textbf{1} & \textbf{2} & \textbf{3} \\
    \midrule
    % Sequential FT-F &  &  & &  \\
    Sequential FT-P & 26.7 & 28.8 & 27.4 & 26.6 \\
    % Per-task Finetune & 70.0  & 70.0  & 70.0  & 70.0 \\
    Per-task FT & 76.6 & 76.6 & 76.6 & 76.6 \\
    % EWC$^\diamond$  & 40.6  & 39.0  & 38.0  & 44.8 \\
    % LFPT5$^\diamond$ & 52.7  & 47.6  & 52.6  & 57.9 \\
    EPI & 48.4 & 48.1 & 48.0  & 49.0 \\
    O-LoRA$^\ddagger$ & 76.1 & 76.8 & 75.7 & 75.7 \\
    % EPI (TIL) & 75.1  & 75.1 & 75.1 & 75.1 \\
    % \textbf{Ours} (CIL) &       &       &       &  \\
    \textbf{\mocl}\ \textbf{(Ours)} & \textbf{78.2} & \textbf{78.4} & \textbf{77.7} & \textbf{78.4} \\
    \bottomrule
    \end{tabular}%
    }
  \caption{TIL results on the far-domain MTL5 dataset with BERT, T5 and Llama 2 as the base model. The superscripts $^\diamond$, $^\dagger$ and $^\ddagger$ indicate that results are taken from \newcite{razdaibiedina2022progressive}, \newcite{wang-etal-2023-rehearsal} and \newcite{wang2023orthogonal}, respectively.\protect\footnotemark[7]}
  \label{tab:res-til-far}%
\end{table*}

\begin{table}[h]
  \centering
  \scalebox{0.71}{
    \begin{tabular}{lcccc}
    \toprule
    \multicolumn{1}{c}{\multirow{2}[2]{*}{\textbf{CIL}}} & \multicolumn{4}{c}{\textbf{Datasets}} \\
          & \multicolumn{1}{c}{\textbf{WOS}} & \textbf{AfriSenti} & \textbf{MTL5-BERT} & \textbf{MTL5-T5} \\
    \midrule
    EPI   & 77.83 & 43.10 & \textbf{77.3}  & 56.4 \\
    \textbf{\mocl}\ \textbf{(Ours)} & \textbf{79.23}  & \textbf{45.62} & 74.1  &  \textbf{56.8} \\
    \bottomrule
    \end{tabular}%
    }
  \caption{CIL results. We only compare \mocl\ and EPI as they are the only two rehearsal-free approaches that support this challenging task setting.}
  \label{tab:res-cil}%
\end{table}%

\begin{table}[h]
  \centering
  \scalebox{0.71}{
    \begin{tabular}{lcccc}
    \toprule
    \multicolumn{1}{c}{\multirow{2}[2]{*}{\textbf{FWT}}} & \multicolumn{4}{c}{\textbf{Datasets}} \\
          & \multicolumn{1}{c}{\textbf{WOS}} & \textbf{AfriSenti} & \textbf{MTL5-BERT} & \textbf{MTL5-T5} \\
    \midrule
    ProgPrompt & 8.4 & -3.5 & -0.3 & 0 \\
    % EPI   &   0    &   0    &   0   &  0 \\
   \textbf{\mocl}\ \textbf{(Ours)}  & \textbf{8.9} &  \textbf{4.8} & \textbf{0.3} & \textbf{0.3} \\
    \bottomrule
    \end{tabular}%
    }
  \caption{Forward transfer (FWT) score comparison between ProgPrompt and \mocl\ across datasets.}
  \label{tab:res-fwt}%
\end{table}%

\subsection{\mocl\ for Class-Incremental Learning}
\label{sec:res-cil}
Table \ref{tab:res-cil} presents the class-incremental results. We compare \mocl\ only to EPI as they are the only two rehearsal-free CL methods applicable to this setting.
Unlike EPI, our model has no explicit task identification component.
Nevertheless, it still achieves better or competitive results. % compared to EPI.

\section{\blue{Analysis}}
\blue{In this section, we analyze \mocl's forward transfer capability and its matching weights distribution.}
\subsection{Forward Transfer Analysis}
\label{sec:res-fwt}

We calculate the forward transfer scores (FWT) \cite{wang2023comprehensive} of \mocl\ and ProgPrompt in the TIL setting (see Table \ref{tab:res-fwt}).\footnote[6]{As mentioned in \ref{sec:res-til}, EPI consists of task identification and per-task FT. Thus, with given task IDs, EPI is identical to per-task FT, thus, includes no knowledge transfer ($\text{FWT}=0$).} \blue{The FWT metric evaluates the average influence of all previous tasks on the current task:}

{\small\begin{equation}
    \text{FWT} = \frac{1}{N - 1} \sum_{j=2}^{N} (a_{i,i} - \tilde{a}_{i}),
\end{equation}}\noindent\blue{where $N$ is the number of tasks in the continual learning sequence, $a_{i,i}$ denotes the performance evaluated on the $i$-th task after incremental learning on the first $i$ tasks, $\tilde{a}_{i}$ is the task performance of a randomly initialized reference model trained with dataset $D_i$.} The results show that ProgPrompt suffers from catastrophic forgetting on AfriSenti ($\text{FWT}<0$) and explain the performance gap in Table \ref{tab:res-til-near}.
We assume the reason is negative interference between some of the languages, as observed in \newcite{wang2023nlnde}.
%, training on some languages makes the model's performance degrade in other languages. 
ProgPrompt suffers from such interference as it concatenates all previous task-specific modules with the current module, without considering task interaction. In contrast, \mocl\ composes task modules via task matching, thus avoiding negative interference between tasks while exploiting similarities for knowledge transfer.

On the far-domain MTL5 dataset, \mocl\ still achieves higher scores than ProgPrompt. This suggests that our approach is better at transferring knowledge on various benchmarks, even with different levels of task similarities.

\subsection{\blue{Task Matching Weights Visualization}}
\label{sec:weight_dist}

\footnotetext[7]{\blue{Among these baseline methods, ProgPrompt is only applicable with prefix-tuning as the PEFT module. O-LoRA is specifically designed for LoRA as the PEFT module.}}

\begin{figure}
    \centering
    \includegraphics[width=\linewidth]{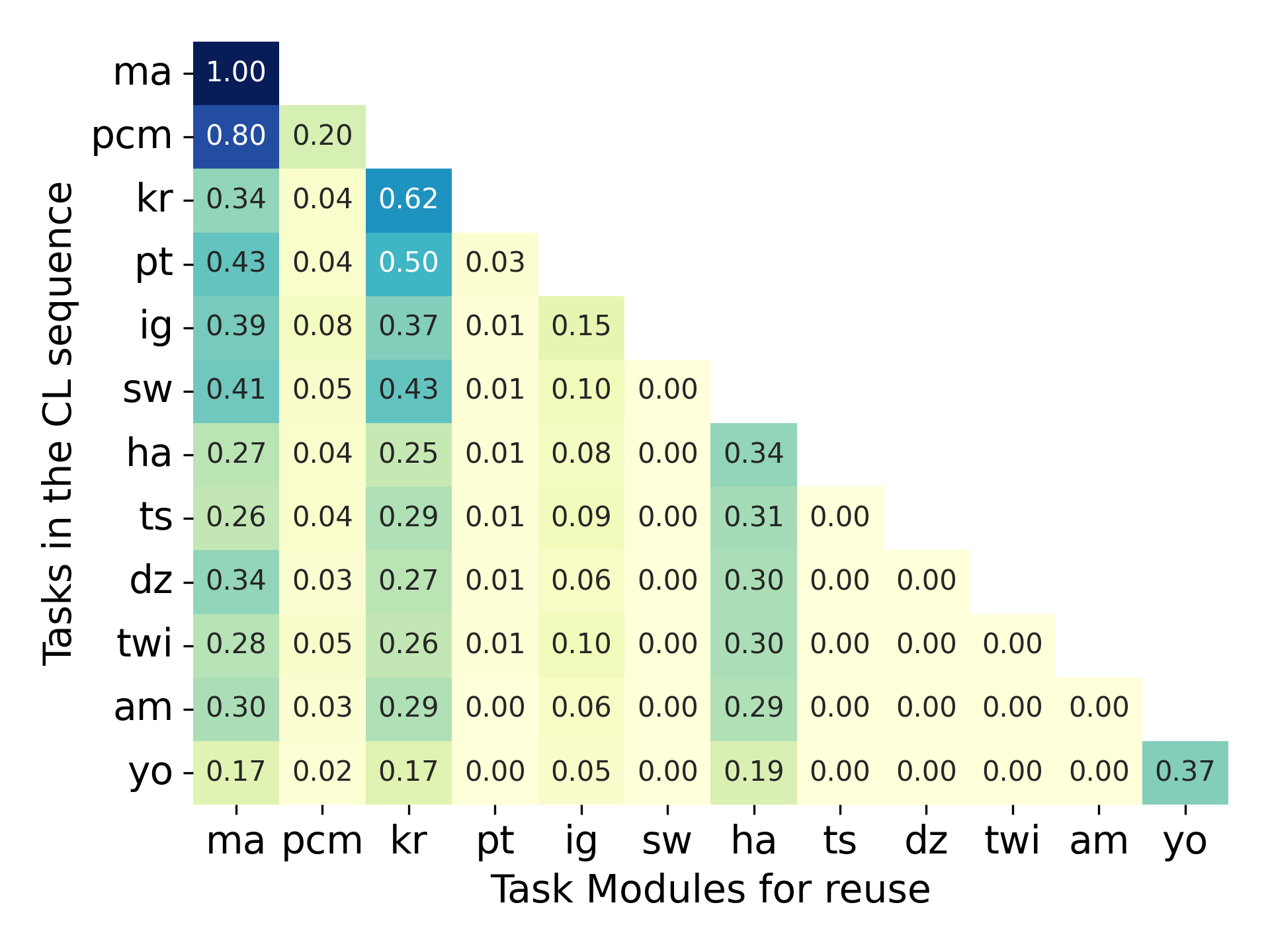}
    \caption{\blue{Visualization on the task matching weights of \mocl\ on the AfriSenti dataset (Task order 2). The heatmap entries quantify the extent of contribution from each task-specific module (denoted on the x-axis) to the subsequent tasks (represented on the y-axis).}}
    \label{fig:weight_dist_afrisenti}
\end{figure}

\blue{In Figure \ref{fig:weight_dist_afrisenti}, we visualize the task matching weight distribution of \mocl\ on the AfriSenti dataset\footnote[8]{We provide weights visualization on other datasets in Appendix~\ref{sec:res-details}} exemplarily with task order 2 (see Table \ref{tab:task-orders}) under the TIL setting. As \mocl\ performs per-instance task matching and module composition, we average the weights across all examples from a given task (i.e., language). As introduced in Section \ref{sec:method}, while learning on the $n^{\text{th}}$ task, we calculate the cosine similarity between the input embeddings and task feature vectors up to the current $n^{\text{th}}$ task. Therefore, the heatmap only has the lower left part. }

\blue{Certain task-specific modules, such as \textit{ma}, \textit{kr}, and \textit{ha}, exhibit utility across a wide range of other tasks, while others, like \textit{pcm}, demonstrate utility exclusivity in their respective tasks. Moreover, we observe that there is a pronounced sparsity in the learned weight distributions. Our task matching paradigm can be considered a mixture-of-experts strategy where we use task-specific experts as the mixture components. Such a sparsity suggests that we can potentially reduce the number of experts, instead of using experts specific to each task. This can be an interesting direction for future work.}

\label{sec:bibtex}

\section{Conclusion}

In this paper, we introduced \mocl, 
 a modular and compositional continual learning framework for language models, effectively addressing the critical challenges of catastrophic forgetting and knowledge transfer in continual learning. 
% In particular, \mocl\ deftly balances the trade-off between knowledge interference and transfer by introducing task feature vectors for module composition. 
Our broad evaluations across various benchmarks demonstrated \mocl's superior performance compared to 
% in not only outperforming 
existing state-of-the-art methods and showed its proficiency in knowledge transfer from previous tasks. %Additionally, the task matching strategy empowers \mocl\ with test-time task inference capabilities, further showing its effectiveness in class-incremental scenarios. % where task identities are unknown during testing. %Overall, \mocl\ is a simple but effective rehearsal-free continual learning framework, offering a practical solution to continual learning of language models in dynamically evolving environments.

\section*{Limitations}
One limitation of our work is the scope of evaluation. While \mocl\ is generally 
% compatible with different types of transformer-based models and 
applicable to a wide range of tasks, we primarily focus on text classification tasks 
% with encoder-based and encoder-decoder-based language models 
following prior work. 
Further experiments with other types of NLP tasks, especially generative tasks is left as a future work direction.

\blue{Besides, the continual learning datasets we study in this work include at most 12 tasks in a sequence. As the continual learning sequence scales to dozens or hundreds of tasks, continually initializing a new PEFT module for each task would largely increase the computational and storage cost. In Section~\ref{sec:weight_dist}, we observe that the learned weight distribution is notably sparse, suggesting that we could reduce the number of task modules instead of assigning a specific module for each task. It would be an interesting future work direction to utilize some module pruning strategies for more efficient continual learning.}
% Besides, \mocl\  leverages prefix-tuning for parameter-efficient continual learning. It has not been evaluated with other prevalent parameter-efficient fine-tuning (PEFT) approaches such as Adapter \cite{houlsby2019parameter} or LoRA \cite{hu2021lora}. Future work could explore the synergy between our method and these alternative fine-tuning strategies.

% Further exploration of the decision making process of module composition, i.e., to determine to what extent previous learned modules can be reused for current task learning.

\section*{Acknowledgements}
We would like to thank the anonymous reviewers for their constructive feedback, which was very valuable for improving our work. This work was partially supported by Deutsche Forschungsgemeinschaft (project SCHU 2246/14-1).

\bibliography{anthology,custom}
\nocite{Ando2005,andrew2007scalable,rasooli-tetrault-2015}

\newpage

\appendix
\section{Appendix}
\label{sec:appendix}
\subsection{Dataset Information}
\label{sec:dataset-info}
Here we give detailed information of the datasets we use with in this work. 
For \textit{near-domain} benchmarks, we use Web-of-Science (WOS) and AfriSenti.  WOS is originally a hierarchical document classification datasets which collects published papers in 7 different domains, which are biochemistry, civil engineering, computer science, electrical engineering, medical science, mechanical engineering and psychology. These domains corresponds to 7 high-level classes for document classification,  and there are several low-level subclasses under each high-level class. Following \newcite{wang-etal-2023-rehearsal}, we organize 7 continual learning tasks according to these high-level classes. AfriSenti is a multilingual sentiment analysis dataset which covers 12 low-resource African languages, including Amharic (am), Algerian Arabic (dz), Hausa (ha), Igbo (ig), Kinyarwanda(kr), Moroccan Arabic (ma), Nigerian Pidgin (pcm), Mozambican Portuguese (pt), Swahili (sw), Xitsonga (ts), Twi (twi) and Yoruba (yo).

For \textit{far-domain} benchmarks, we adopt the commonly used MTL5 dataset, consisting of 5 text classification tasks. \blue{Detailed task information is given in Table \ref{tab:mtl5-dataset}}. We experiment with BERT-base and T5-large models on this dataset in line with prior work \cite{razdaibiedina2022progressive}. For BERT-based experiments, we uses the same train and test sets following prior work such as ProgPrompt \cite{razdaibiedina2022progressive} and EPI \cite{wang-etal-2023-rehearsal}, consisting of 115,000 training and 7,600 text samples for each task. For T5- and Llama 2-based experiments, 4 out of these 5 tasks (except Yelp) are used in line with \newcite{razdaibiedina2022progressive} and \cite{wang2023orthogonal}, with 16 samples per task for training and the test sets are unchanged.

Following prior work, we report F1 score on the AfriSenti dataset \cite{muhammad2023afrisenti, wang-etal-2023-gradsim} and accuracy on WOS and MTL5 datasets \cite{de2019episodic, razdaibiedina2022progressive, wang-etal-2023-rehearsal}.
We use different task orders for each dataset to evaluate the robustness of continual learning methods against changing task orders. The task orders used are summarzied in Table \ref{tab:task-orders}.

\begin{table*}[htbp]
  \centering
  \scalebox{0.9}{
    \begin{tabular}{cccl}
    \toprule
    \textbf{Dataset} & \textbf{Order} & \textbf{Model} & \textbf{Task Sequence} \\
    \midrule
    \multirow{3}[2]{*}{AfriSenti} & 1     & AfroXLMR &  am $\rightarrow$ dz $\rightarrow$ ha $\rightarrow$ ig $\rightarrow$ kr $\rightarrow$ ma $\rightarrow$ pcm $\rightarrow$ pt $\rightarrow$ sw $\rightarrow$ ts $\rightarrow$ twi $\rightarrow$ yo \\
          & 2     & AfroXLMR & ma $\rightarrow$ pcm $\rightarrow$ kr $\rightarrow$ pt $\rightarrow$ ig $\rightarrow$ sw $\rightarrow$ ha $\rightarrow$ ts $\rightarrow$ dz $\rightarrow$ twi $\rightarrow$ am $\rightarrow$ yo \\
          & 3     & AfroXLMR & am $\rightarrow$ dz $\rightarrow$ ha $\rightarrow$ ma $\rightarrow$ ig $\rightarrow$ kr $\rightarrow$ sw $\rightarrow$ ts $\rightarrow$ twi $\rightarrow$ yo $\rightarrow$ pcm $\rightarrow$ pt \\
    \midrule
    WOS   & 1     & BERT  & 1 $\rightarrow$ 2 $\rightarrow$ 3 $\rightarrow$ 4 $\rightarrow$ 5 $\rightarrow$ 6 $\rightarrow$ 7 \\
    \midrule
    \multirow{4}[2]{*}{MTL5} & 1     & BERT  & ag $\rightarrow$ yelp $\rightarrow$ amazon $\rightarrow$ yahoo $\rightarrow$ db \\
          & 2     & BERT  & yelp $\rightarrow$ yahoo $\rightarrow$ amazon $\rightarrow$ db $\rightarrow$ agnews \\
          & 3     & BERT  & db $\rightarrow$ yahoo $\rightarrow$ ag $\rightarrow$ amazon $\rightarrow$ yelp \\
          & 4     & BERT  & yelp $\rightarrow$ ag $\rightarrow$ db $\rightarrow$ amazon $\rightarrow$ yahoo \\
    \midrule
    \multirow{3}[2]{*}{MTL5} & 1     & T5, Llama 2    & db $\rightarrow$ amazon $\rightarrow$ yahoo $\rightarrow$ ag \\
          & 2     & T5, Llama 2    & db $\rightarrow$ amazon $\rightarrow$ ag $\rightarrow$ yahoo \\
          & 3     & T5, Llama 2    & yahoo $\rightarrow$ amazon $\rightarrow$ ag $\rightarrow$ db \\
    \bottomrule
    \end{tabular}%
    }
  \caption{The different orders of task sequences used for continual learning experiments.}
  \label{tab:task-orders}%
\end{table*}%
\subsection{Continual Learning Setting Details}
\label{sec:cl-setting-details}
Beyond the general formulation as introduced in Section \ref{sec:cl-basics}, continual learning can be categorized into several detailed settings,\footnote[9]{We focus on some commonly studied continual learning settings here, for a more comprehensive categorization of continual learning settings please refer to \cite{wang2023comprehensive}.} according to the distinction between incremental data batches and task identity availability.
\textit{Task-incremental learning}  (TIL) refers to the scenario where the tasks have disjoint label space. Task identities are provided in both training and testing. This is the most studied continual learning scenario and also the easiest case of continual learning tasks.

\textit{Class-incremental learning} (CIL) is a more challenging continual learning scenario where the task identities are not available during testing. The tasks still have disjoint label space and task identities are available during training.

\textit{Domain-incremental learning} (DIL) assumes the class labels are the same across all tasks and the inputs are from different domains. Whether task identities are given during testing or not, it all belongs to this category. Strictly speaking, the AfriSenti benchmark used in this work belongs to the DIL category. In this multilingual sentiment analysis dataset, the data of different tasks (languages) is considered to have different input distributions, while the label space is shared across tasks (languages). In this work, we aim to evaluate \mocl\ in settings where the task identities are provided and are not provided during testing. We also consider the evaluation setting on AfriSenti as task-incremental learning and class-incremental learning, respectively. In our experiments, we assume tasks have disjoint label spaces, i.e., their classification heads are different. In this way, we use the AfriSenti benchmark for TIL and CIL evaluation as well.

\subsection{Experimental Setup Details}
In this section, we give more detailed information about the baseline methods we used in this work and the implementation details for experiments.

\begin{table}[htbp]
  \centering
  \scalebox{0.7}{
    \begin{tabular}{l|cccc}
    \toprule
    \textbf{Method} & \textbf{RF} & \textbf{PE} & \textbf{CI} & \textbf{KT} \\
    \midrule
    EWC \cite{kirkpatrick2017overcoming} & \checkmark     &       &       & \checkmark \\
    MBPA++ \cite{de2019episodic} &       &       & \checkmark     & \checkmark \\
    IDBR \cite{huang-etal-2021-continual} &       &       & \checkmark     & \checkmark \\
    LFPT5 \cite{qin2021lfpt5}&       & \checkmark     &       & \checkmark \\
    ProgPrompt \cite{razdaibiedina2022progressive} & \checkmark     & \checkmark     &       & \checkmark \\
    EPI \cite{wang-etal-2023-rehearsal} & \checkmark     & \checkmark     & \checkmark     &  \\
    O-LoRA \cite{wang2023orthogonal} & \checkmark     & \checkmark     & \checkmark     &  \\
    \textbf{\mocl\ (Ours)} & \checkmark     & \checkmark     & \checkmark     & \checkmark \\
    \bottomrule
    \end{tabular}%
  }
  \caption{Comparison between \mocl\ and existing CL approaches. RF: rehearsal-free; PE: parameter-efficient; CI: applicable to class-incremental learning, KT: enabled knowledge transfer.}
  %Specifically, RF indicates whether the approach is rehearsal-free; PE indicates whether the CL process is parameter-efficient; CI means whether the approach is applicable for CIL, where the task identities are not available during testing; KT indicates whether knowledge transfer is enabled during continual learning.}
  \label{tab:addlabel}%
\end{table}%

\subsubsection{Baseline Methods}
\label{sec:baseline-details}
In Sections \ref{sec:exp-res} and \ref{sec:res-all-baselines}, we evaluate \mocl\ and prior continual learning methods on different benchmark datasets. Here we give a more detailed description of the baseline methods used in this work.
% Replay: saving part of the data samples from previous tasks and alternately training the model on the memory and the new task data.
 
% EWC: a commonly used regularization-based CL approach which fine-tunes the model with a regularization loss to restrict the parameter change while learning new tasks. 
 
% LwF: a regularization-based CL approach based on knowledge distillation. It considers the old model as the teacher and the updated model as the student and calculates the distillation loss for model fine-tuning. 

\blue{
MBPA++ \cite{de2019episodic}: introduces an episodic memory model that performs sparse experience replay and local adaptation to continuously learn and reuse previously acquired knowledge.
}

\blue{
IDBR \cite{huang-etal-2021-continual}: disentangles the text embeddings into task-generic
space and task-specific space and further regularizes them differently. It also leverages data replay and two auxiliary tasks for effective continual learning. Due to its architectural design, it is only applicable to encoder-based transformer models for classification tasks.
}

\blue{
LFPT5 \cite{qin2021lfpt5}: a continual learning approach based on the T5 model. It continuously trains a soft prompt to solve the task and generate pseudo samples for data replay.   
}

ProgPrompt \cite{razdaibiedina2022progressive}: a parameter isolation-based continual learning method which assigns task-specific parameters to avoid catastrophic forgetting. During continual learning, ProgPrompt progressively concatenates all task-specific modules to encourage forward transfer. Task identities are always required during training and testing.
 
EPI \cite{wang-etal-2023-rehearsal}: a parameter isolation-based method applicable to the class-incremental learning setting. EPI introduces a non-parametric task identification module that identifies tasks during testing. Given reliable task identification, the CIL performance could be comparable with TIL, where the ground truth task identities are given.

\blue{O-LoRA \cite{wang2023orthogonal}: a parameter isolation-based method which learns tasks in different low-rank vector spaces that are kept orthogonal to each other to minimize interference. It mitigates catastrophic forgetting by constraining the gradient update of the current task to be orthogonal to the gradient space of the past tasks. However, the orthogonality of the gradient subspace for individual tasks also limits knowledge transfer between tasks.}

As discussed in the main paper, ProgPrompt and EPI are two closely related prior work to \mocl. 
% We illustrate their difference in Figure \ref{fig:}. 
ProgPrompt concatenates all previously learned parameters with the current learnable to encourage knowledge transfer while ignoring different levels of relatedness across tasks: There might be knowledge interference or transfer between different pairs of tasks. EPI focus on the class-incremental learning setting and the task-specific parameters are completely isolated, i.e., there is no knowledge transfer in their approach. In contrast, \mocl\ assigns different weights to previously learned task-specific modules based on the relatedness between tasks, therefore deftly balancing knowledge interference or transfer and leading to more effective knowledge transfer.

\subsection{Additional Experimental Results}
\label{sec:res-details}
In this section, we give additional experimental resu\textit{}lts, including \blue{the additional baseline results}, \mocl's per-task results on the three datasets, and the weight distribution on AfriSenti for module composition.

\begin{table}[htbp]
  \centering
  \scalebox{0.85}{
    \begin{tabular}{l|ccccc}
    \toprule
    \multicolumn{1}{c|}{\multirow{2}[2]{*}{\textbf{Method}}} & \multicolumn{5}{c}{\textbf{MTL5 (BERT) Orders}} \\
          & \textbf{AVG} & \textbf{1} & \textbf{2} & \textbf{3} & \textbf{4} \\
    \midrule
    \blue{Sequential FT-F} & \blue{14.8}  & \blue{27.8}  & \blue{26.7}  & \blue{4.5}   & \blue{18.4} \\
    Sequential FT-P & 66.5  & 66.4  & 65.7  & 65.4   & 68.5 \\
    Per-task FT & 79.0  & 79.0  & 79.0  & 79.0  & 79.0 \\
    \blue{\textit{MBPA++*}} & \blue{\textit{70.6}}  & \blue{\textit{70.8}}  & \blue{\textit{70.9}}  & \blue{\textit{70.2}}  & \blue{\textit{70.7}} \\
    \blue{\textit{IDBR*}} & \blue{\textit{76.3}}  & \blue{\textit{75.9}}  & \blue{\textit{76.2}}  & \blue{\textit{76.4}}  & \blue{\textit{76.7}} \\
    ProgPrompt* & 77.9  & 78.0  & 77.9  & 77.9  & 77.9 \\
    EPI* & 77.3  & 77.4  & 77.3  & 77.2  & 77.4 \\
    \textbf{\mocl}\ \textbf{(Ours)} & \textbf{79.4} & \textbf{79.3} & \textbf{79.6} & \textbf{79.2} & \textbf{79.4} \\
    \bottomrule
    \end{tabular}}
  \scalebox{0.85}{
    \begin{tabular}{l|cccc}
    \toprule
    \multicolumn{1}{c|}{\multirow{2}[2]{*}{\textbf{Method}}} & \multicolumn{4}{c}{\textbf{MTL5 (T5) Orders}} \\
          & \textbf{AVG} & \textbf{1} & \textbf{2} & \textbf{3} \\
    \midrule
    \blue{Sequential FT-F} & \blue{28.5}  & \blue{18.9}  & \blue{24.9}  & \blue{41.7} \\
    Sequential FT-P & 27.2  & 24.6  & 30.3 & 25.0 \\
    Per-task FT & 75.1  & 75.1 & 75.1 & 75.1 \\
    \blue{\textit{LFPT5*}} & \blue{\textit{52.7}}  & \blue{\textit{47.6}}  & \blue{\textit{52.6}}  & \blue{\textit{57.9}} \\
    ProgPrompt* & 75.1  & 75.0  & 75.0  & 75.1 \\
    EPI & 56.4 & 49.7 & 54.1 & 65.3\\
    \textbf{\mocl}\ \textbf{(Ours)} & \textbf{75.9} & \textbf{75.6} & \textbf{75.4} & \textbf{76.7} \\
    \bottomrule
    \end{tabular}%
    }
  \caption{\blue{TIL results with additional baseline methods on far-domain MTL5 with BERT and T5 as the base model. * indicates that results are taken from the corresponding papers. \mocl\ still outperforms all existing work in both evaluation settings.}}
  \label{tab:res-til-far-all-baselines}%
\end{table}%

\noindent \blue{\textbf{Additional baselines.}}
\label{sec:res-all-baselines}
\blue{
In Section \ref{sec:baselines} and \ref{sec:exp-res}, we only include methods that are applicable across models for consistency reasons. In Table \ref{tab:res-til-far-all-baselines}, we provide results with three additional continual learning methods, where IDBR \cite{huang-etal-2021-continual} and MBPA++ \cite{de2019episodic} are BERT-based continual learning methods, while LFPT5 \cite{qin2021lfpt5} is specifically designed for the T5 language model. In both evaluation settings, \mocl\ consistently shows better performance than prior work, demonstrating the effectiveness of our proposed approach.
}

\noindent \textbf{Per-task results.}
From Table \ref{tab:wos-detailed-res} to \ref{tab:mtl5-t5-detailed-res}, we give the detailed per-task results on the aforementioned datasets under task-incremental learning and class-incremental learning settings.

\begin{table}[htbp]
    \footnotesize
  \centering
  \scalebox{0.73}{
    \begin{tabular}{ccccccccc}
    \toprule
    \multicolumn{9}{c}{\textbf{WOS per-task results}} \\
    \midrule
    \textit{\textbf{order 1}} & \textbf{AVG} & \textbf{1} & \textbf{2} & \textbf{3} & \textbf{4} & \textbf{5} & \textbf{6} & \textbf{7} \\
    \midrule
    \textbf{TIL} & 90.59 & 91.86 & 95.72 & 80.05 & 93.25 & 95.09 & 93.60 & 84.54 \\
    \textbf{CIL} & 79.23 & 70.57 & 93.36 & 58.74 & 86.67 & 91.29 & 87.82 & 66.19 \\
    \bottomrule
    \end{tabular}%
    }
  \caption{Detailed per-task results on the WOS dataset under TIL and CIL settings.}
  \label{tab:wos-detailed-res}%
\end{table}%

\begin{table}[htbp]
  \footnotesize
  \centering
  \scalebox{0.85}{
    \begin{tabular}{cccccccc}
    \toprule
    \multicolumn{8}{c}{\textbf{AfriSenti per-task results}} \\
    \midrule
    \textit{\textbf{order1}} & \textbf{AVG} & \textbf{am} & \textbf{dz} & \textbf{ha} & \textbf{ig} & \textbf{kr} & \textbf{ma} \\
    \textbf{TIL} & 57.05 & 58.52 & 58.58 & 66.83 & 56.92 & 63.68 & 48.68 \\
    \textbf{CIL} & 45.57 & 63.56 & 52.88 & 47.06 & 26.15 & 52.16 & 40.28 \\
    \textit{\textbf{order1}} &       & \textbf{pcm} & \textbf{pt} & \textbf{sw} & \textbf{ts} & \textbf{twi} & \textbf{yo} \\
    \textbf{TIL} &       & 60.59 & 64.27 & 57.24 & 42.97 & 46.56 & 59.77 \\
    \textbf{CIL} &       & 56.98 & 36.71 & 28.80 & 38.10 & 44.21 & 60.00 \\
    \midrule
    \textit{\textbf{order2}} & \textbf{AVG} & \textbf{ma} & \textbf{pcm} & \textbf{kr} & \textbf{pt} & \textbf{ig} & \textbf{sw} \\
    \textbf{TIL} & 56.52 & 47.41 & 58.51 & 65.15 & 61.38 & 54.47 & 55.19 \\
    \textbf{CIL} & 44.32 & 40.56 & 57.12 & 47.53 & 35.22 & 25.44 & 29.21 \\
    \textit{\textbf{order2}} &       & \textbf{ha} & \textbf{ts} & \textbf{dz} & \textbf{twi} & \textbf{am} & \textbf{yo} \\
    \textbf{TIL} &       & 67.27 & 44.45 & 61.20 & 45.40 & 58.32 & 59.53 \\
    \textbf{CIL} &       & 44.49 & 40.33 & 46.24 & 41.82 & 64.91 & 59.03 \\
    \midrule
    \textit{\textbf{order3}} & \textbf{AVG} & \textbf{am} & \textbf{dz} & \textbf{ha} & \textbf{ma} & \textbf{ig} & \textbf{kr} \\
    \textbf{TIL} & 56.74 & 58.52 & 58.58 & 66.83 & 50.05 & 54.20 & 59.90 \\
    \textbf{CIL} & 46.95 & 46.00 & 39.34 & 57.76 & 45.17 & 47.08 & 49.89 \\
    \textit{\textbf{order3}} &       & \textbf{sw} & \textbf{ts} & \textbf{twi} & \textbf{yo} & \textbf{pcm} & \textbf{pt} \\
    \textbf{TIL} &       & 57.47 & 42.60 & 44.83 & 60.01 & 60.17 & 64.71 \\
    \textbf{CIL} &       & 53.56 & 23.24 & 34.61 & 49.19 & 53.50 & \textbf{CIL} \\
    \bottomrule
    \end{tabular}%
    }
  \caption{Detailed per-task results on the AfriSenti dataset under TIL and CIL settings.}
  \label{tab:afrisenti-detailed-res}%
\end{table}%

\begin{table}[htbp]
  \footnotesize
  \centering
  \scalebox{0.8}{
    \begin{tabular}{lcccccc}
    \toprule
    \multicolumn{7}{c}{\textbf{MTL5-BERT per-task results}} \\
    \midrule
    \textit{\textbf{order1}} & \textbf{AVG} & \textbf{agnews} & \textbf{yelp} & \textbf{amazon} & \textbf{yahoo} & \textbf{db} \\
    \midrule
    \textbf{TIL} & 79.31 & 94.13 & 64.41 & 61.67 & 77.14 & 99.19 \\
    \textbf{CIL} & 73.02 & 93.39 & 62.75 & 39.13 & 72.30 & 97.52 \\
    \midrule
    \textit{\textbf{order2}} & \textbf{AVG} & \textbf{yelp} & \textbf{amazon} & \textbf{yahoo} & \textbf{db} & \textbf{agnews} \\
    \midrule
    \textbf{TIL} & 79.64 & 64.43 & 62.50 & 78.03 & 99.23 & 94.03 \\
    \textbf{CIL} & 74.00 & 62.69 & 44.91 & 70.98 & 99.14 & 92.26 \\
    \midrule
    \textit{\textbf{order3}} & \textbf{AVG} & \textbf{db} & \textbf{yahoo} & \textbf{agnews} & \textbf{amazon} & \textbf{yelp} \\
    \midrule
    \textbf{TIL} & 79.20 & 99.23 & 77.72 & 94.03 & 61.78 & 63.24 \\
    \textbf{CIL} & 74.75 & 98.40 & 72.19 & 92.97 & 53.82 & 59.57 \\
    \midrule
    \textit{\textbf{order4}} & \textbf{AVG} & \textbf{yelp} & \textbf{agnews} & \textbf{db} & \textbf{amazon} & \textbf{yahoo} \\
    \midrule
    \textbf{TIL} & 79.61 & 64.43 & 94.37 & 99.20 & 62.04 & 77.99 \\
    \textbf{CIL} & 73.55 & 62.54 & 93.41 & 98.98 & 47.75 & 65.07 \\
    \bottomrule
    \end{tabular}%
    }
  \caption{Detailed per-task results on the MTL5 dataset using BERT as the base language model under TIL and CIL settings.}
  \label{tab:mtl5-bert-detailed-res}%
\end{table}%

\begin{table}[htbp]
  \footnotesize
  \centering
  \scalebox{0.95}{
    \begin{tabular}{lccccc}
    \toprule
    \multicolumn{6}{c}{\textbf{MTL5-T5 per-task results}} \\
    \midrule
    \textit{\textbf{order1}} & \textbf{AVG} & \textbf{db} & \textbf{amazon} & \textbf{yahoo} & \textbf{agnews} \\
    \midrule
    \textbf{TIL} & 75.59 & 98.27 & 47.88 & 70.84 & 85.31 \\
    \textbf{CIL} & 51.15 & 40.86 & 11.34 & 67.58 & 84.84 \\
    \midrule
    \textit{\textbf{order2}} & \textbf{AVG} & \textbf{db} & \textbf{amazon} & \textbf{agnews} & \textbf{yahoo} \\
    \midrule
    \textbf{TIL} & 75.37 & 98.18 & 47.99 & 84.69 & 70.64 \\
    \textbf{CIL} & 47.84 & 32.04 & 8.91  & 79.84 & 70.59 \\
    \midrule
    \textit{\textbf{order3}} & \textbf{AVG} & \textbf{yahoo} & \textbf{amazon} & \textbf{agnews} & \textbf{db} \\
    \midrule
    \textbf{TIL} & 76.70 & 71.42 & 51.09 & 86.25 & 97.99 \\
    \textbf{CIL} & 71.47 & 67.75 & 48.37 & 73.92 & 95.82 \\
    \bottomrule
    \end{tabular}%
    }
  \caption{Detailed per-task results on the MTL5 dataset using T5 as the base language model under TIL and CIL settings.}
  \label{tab:mtl5-t5-detailed-res}%
\end{table}%

\begin{figure}[h]
    \centering
    \includegraphics[width=1\linewidth]{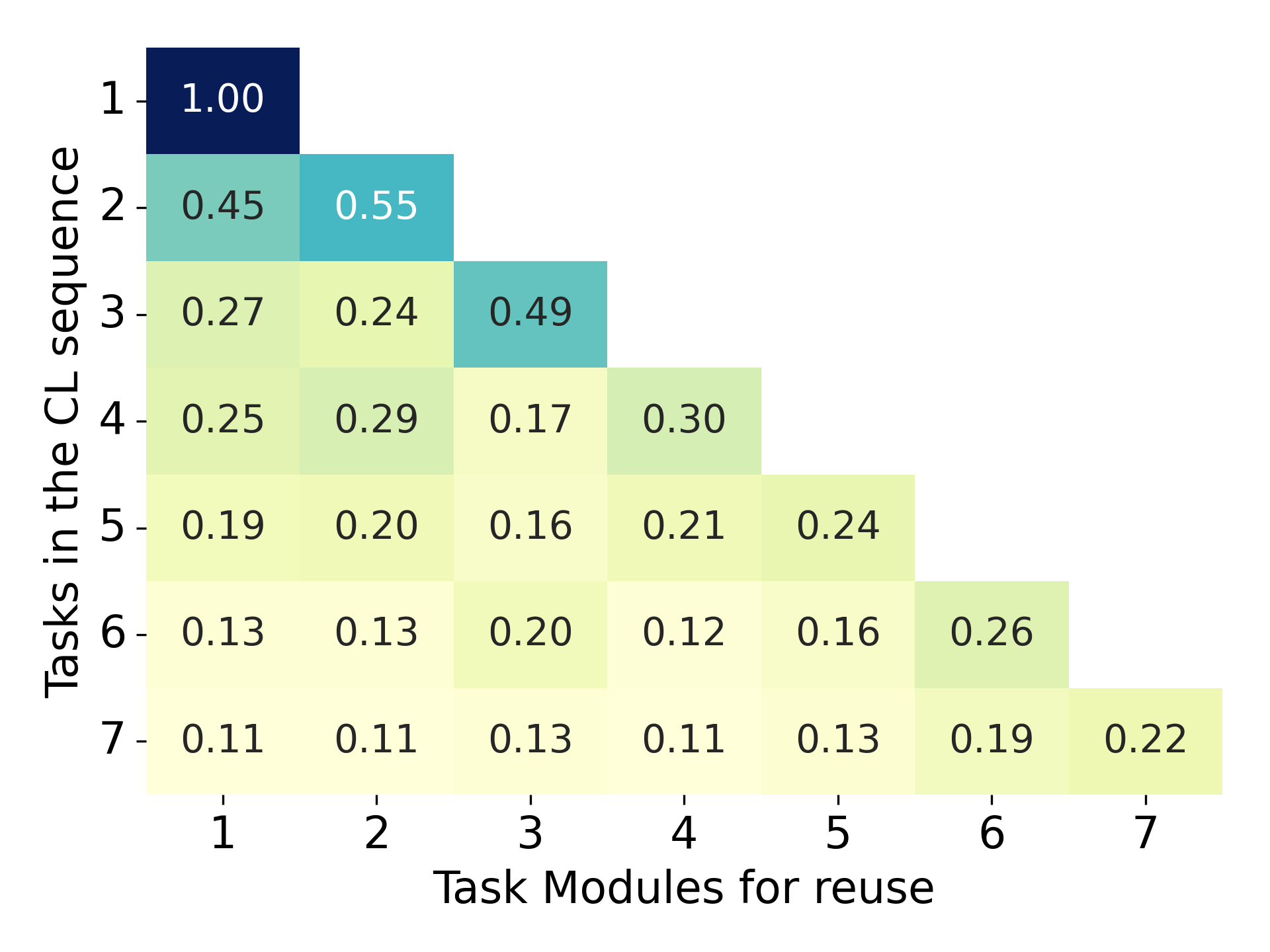}
    \caption{\blue{Visualization on the task matching weights of \mocl\ on the WOS dataset.}}
    \label{fig:weight_dist_wos}
\end{figure}
\begin{figure}[h]
    \centering
    \includegraphics[width=1\linewidth]{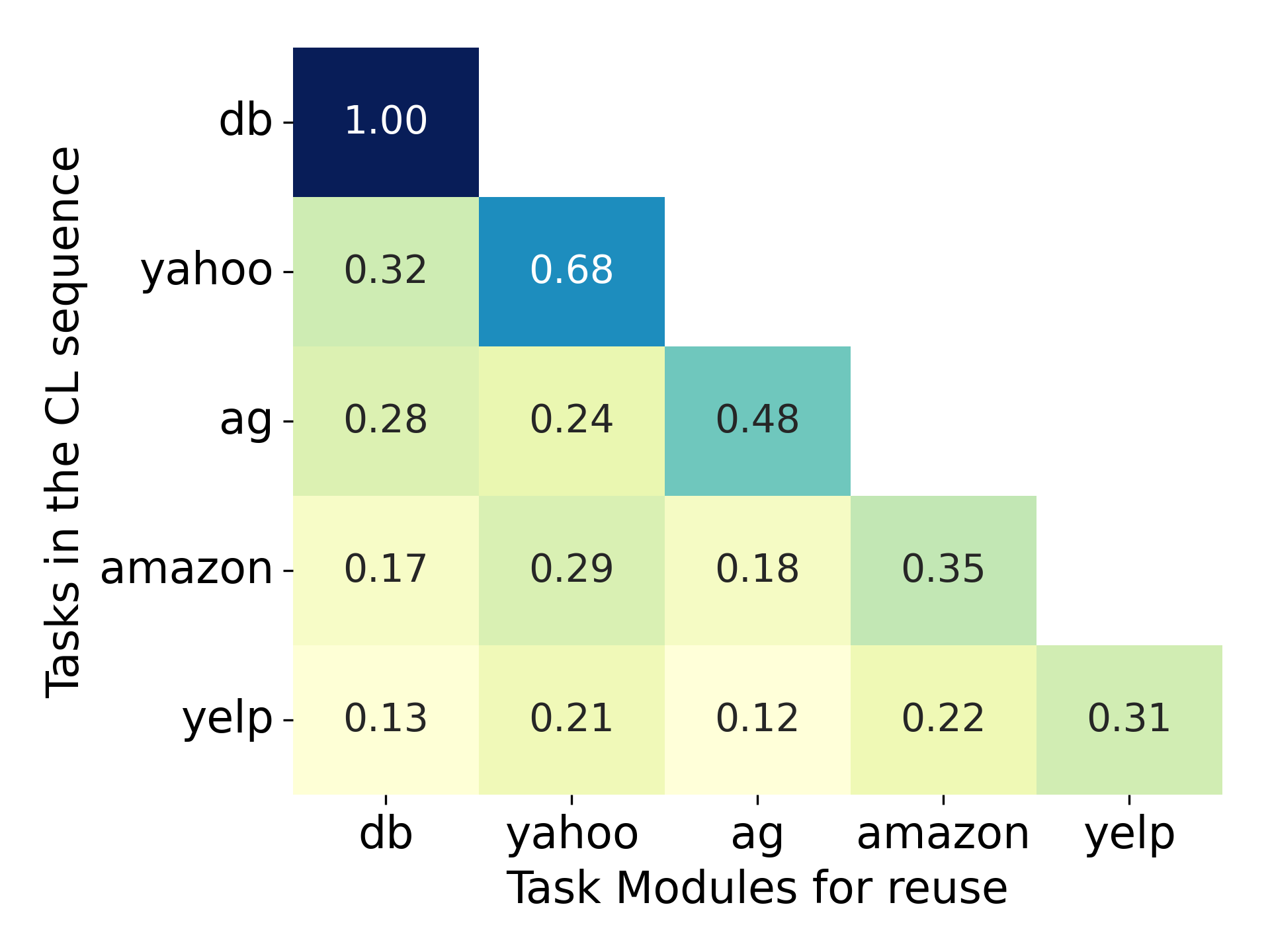}
    \caption{\blue{Visualization on the task matching weights of \mocl\ on the MTL5 dataset with BERT as the base model (Task order 3).}}
    \label{fig:weight_dist_mtl5_bert}
\end{figure}
\begin{figure}[h]
    \centering
    \includegraphics[width=1\linewidth]{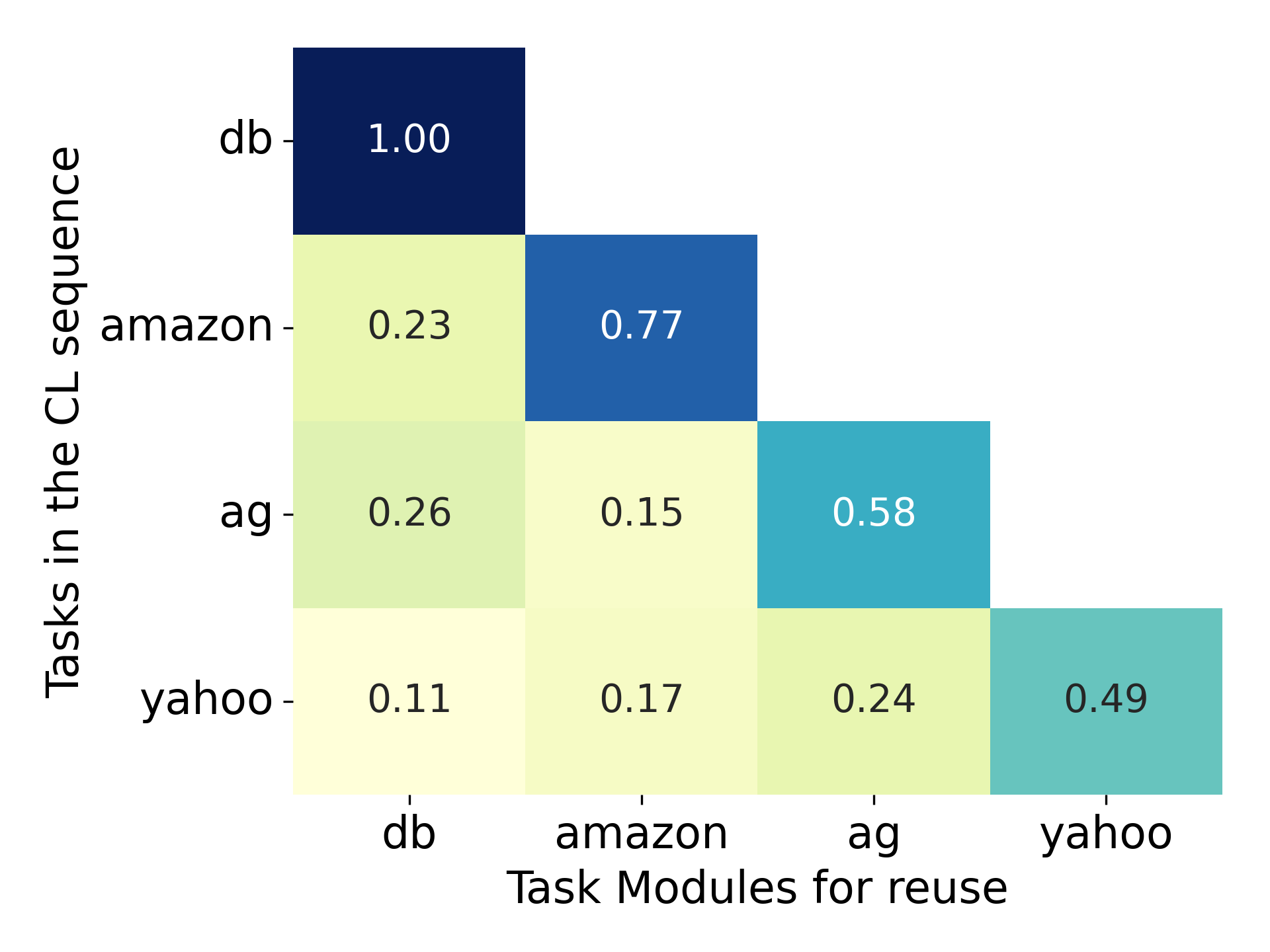}
    \caption{\blue{Visualization on the task matching weights of \mocl\ on the MTL5 dataset with T5 as the base model (Task order 2).}}
    \label{fig:weight_dist_mtl5_t5}
\end{figure}
\noindent \textbf{Task matching weights visualization.}
\blue{
In Section \ref{sec:weight_dist}, we visualized the task matching weights produced by \mocl\ on the AfriSenti dataset (Figure \ref{fig:weight_dist_afrisenti}). In Figures \ref{fig:weight_dist_wos}, \ref{fig:weight_dist_mtl5_bert} and \ref{fig:weight_dist_mtl5_t5}, we provide the visualization on the other datasets. We randomly pick one task order for each dataset for space reasons. As described in Section \ref{sec:weight_dist}, the heatmap entries quantify the extent of contribution from each task-specific module (denoted on the x-axis) to the subsequent tasks (represented on the y-axis).
}

\blue{
We observe a different distribution of weights on the two types of benchmarks, i.e., near-domain and far-domain. On the near-domain datasets, i.e., AfriSenti and WOS, as shown in Figure \ref{fig:weight_dist_afrisenti} and \ref{fig:weight_dist_wos}, the subsequent tasks tend to reuse modules of previous tasks. Whereas on the far-domain MTL5 dataset, Figure \ref{fig:weight_dist_mtl5_bert} and \ref{fig:weight_dist_mtl5_t5} show that the task-specific modules always have higher weights, i.e., the highest values on the diagonal of the heatmap. This is related to the nature of these benchmarks: The tasks from the near-domain benchmark are more related to each other, so there is a tendency for the tasks to reuse existing knowledge from previous modules. While the tasks from the far-domain dataset are dissimilar, and thus the task-specific modules have higher weights.
}

\subsubsection{Implementation Details}
\label{sec:implement-details}
We use the AdamW optimizer \cite{adamw} and the batch size of 8 for all experiments. We choose the same maximum sequence length and prefix length as prior work \cite{razdaibiedina2022progressive, wang-etal-2023-rehearsal}. Table \ref{tab:hyperparams} gives detailed hyperparameter choices of \mocl\ across different datasets. The training
was performed on Nvidia A100 GPUs.\footnote[10]{All experiments ran on a carbon-neutral GPU cluster.}

\begin{table}[h]
  \centering
    \begin{tabular}{lc}
    \toprule
    \textbf{Hyperparameters} &  \\
    \midrule
    \multicolumn{2}{c}{\textit{WOS-BERT}}\\
    \midrule
    Epochs & 40 \\
    Early stop patience & 5 \\
    Learning rate & 3e-2 \\
    Max. sequence len. & 256 \\
    Prefix len. & 16 \\
    \midrule
    \multicolumn{2}{c}{\textit{AfriSenti-AfroXLMR}} \\
    \midrule
    Epochs & 40 \\
    Early stop patience & 5 \\
    Learning rate & 2e-4 \\
    Max. sequence len. & 128 \\
    Prefix len. & 8 \\
    \midrule
    \multicolumn{2}{c}{\textit{MTL5-BERT}} \\
    \midrule
    Epochs & 40 \\
    Early stop patience & 5 \\
    Learning rate & \makecell{8e-4 (db), 1e-3 (yahoo)\\2e-3 (others)} \\
    Max. sequence len. & 256 \\
    Prefix len. & 20 \\
    \midrule
    \multicolumn{2}{c}{\textit{MTL5-T5}} \\
    \midrule
    Epochs & 40 \\
    Early stop patience & 5 \\
    Learning rate &  \makecell{2e-2 (yahoo, db)\\5e-2 (others)} \\
    Max. sequence len. & 512 \\
    Prefix len. & 50 \\
    \midrule
    \multicolumn{2}{c}{\textit{MTL5-Llama 2}} \\
    \midrule
    Epochs & 40 \\
    Early stop patience & 5 \\
    Learning rate &  \makecell{1e-3} \\
    Max. sequence len. & 512 \\
    LoRA rank & 4 \\
    \bottomrule
    \end{tabular}%
  \caption{Hyperparameters used in this work across different CL experiments.}
  \label{tab:hyperparams}%
\end{table}%

% \paragraph{Task matching accuracy}

% \begin{table}[htbp]
%   \centering
%   \footnotesize
%   \caption{Add caption}
%     \begin{tabular}{lccc}
%     \toprule
%     \textbf{Method} & \makecell{\textbf{Average} \\ \textbf{matching acc}} & \textbf{CIL results} & \textbf{TIL results} \\
%     \midrule
%     \multicolumn{4}{c}{\textit{WOS-BERT}} \\
%     \midrule
%     MCCL  &       & 79.20 & 90.59 \\
%     EPI   &       & 77.80 & 82.78 \\
%     \midrule
%     \multicolumn{4}{c}{\textit{AfriSenti-AfroXLMR}} \\
%     \midrule
%     MCCL  &       & 45.62 & 56.77 \\
%     EPI   &       & 43.10 & 52.41 \\
%     \midrule
%     \multicolumn{4}{c}{\textit{MTL5-BERT}} \\
%     \midrule
%     MCCL  &       & 74.10 & 79.40 \\
%     EPI   &       & 77.30 & 79.00 \\
%     \midrule
%     \multicolumn{4}{c}{\textit{MTL5-T5}} \\
%     \midrule
%     MCCL  &       & 56.80 & 75.90 \\
%     EPI   &       & 56.40 & 75.10 \\
%     \bottomrule
%     \end{tabular}%
%   \label{tab:addlabel}%
% \end{table}%
\end{document}